\documentclass[letterpaper,conference]{IEEEtran}
\usepackage{color}
\usepackage{epsfig}
\usepackage{amsmath}
\usepackage{amsfonts}
\usepackage{amsthm}
\usepackage{amssymb}
\usepackage{graphics}
\usepackage{graphicx}
\usepackage{float}
\usepackage{algorithm}
\usepackage{algorithmic}
\usepackage{svg}
\usepackage {epstopdf}
\usepackage{flushend}
\usepackage{bbm}

\title{\LARGE \bf Minimizing the Outage Probability in a Markov Decision Process}

\author{Vincent Corlay and Jean-Christophe Sibel \\
\small{Mitsubishi Electric R\&D Centre Europe, Rennes, France. E-mail: \{v.corlay, \ j.sibel\}@fr.merce.mee.com.} }

\begin{document}

\maketitle

\begin{abstract}
Standard Markov decision process (MDP) and reinforcement learning algorithms optimize the policy with respect to the expected gain.
We propose an algorithm which enables to optimize an alternative objective: the probability that the gain is greater than a given value. The algorithm can be seen as an extension of the value iteration algorithm.
We also show how the proposed algorithm could be generalized to use neural networks, similarly to the deep $Q$ learning extension of $Q$ learning.
\end{abstract}


\section{Introduction}

We consider an agent trying to learn the optimal sequence of actions to be taken in an environment in order to maximize the total amount of reward. 
This total amount of reward is called the gain.
Standard approaches, both in the framework of MDP and reinforcement learning, focus on the expected (average) gain. Consequently, most algorithms in the literature are designed to maximize this expected gain. 
For instance, all algorithms in the reference books \cite{Puterman1994} and \cite{Sutton2020} optimize the expected gain. 
See Section~\ref{sec_review_recent_work} for connections with recent works.
As an illustration in the scope of a resource allocation problem, a scheduler (the agent in this case) should decide which entity gets the resource at each time step, see \cite{Sibel2023} for an example. In the standard approach, the guideline is to minimize the average number of system failures.

However, one may be interested in ensuring that the gain is almost always greater than a given value, i.e., optimizing the outage probability. This is relevant for safety applications where some situations should be avoided with the highest possible probability, even if the average performance is reduced. 
In the case of a resource allocation problem, it can be relevant to minimize the probability that the number of system failures is greater than a given quantity. In information theory, the outage probability of a communication channel, defined as the probability that a given information rate is not supported, is an important metric.

Consequently, we propose an algorithm, inspired from existing MDP algorithms, which enables to optimize the outage probability.
We also explain how this algorithm can be generalized to allow the use of neural networks, instead of storing many values in a table.

\section{Standard maximum expected gain approach}
\label{sec_standard_expe_gain_policy}

We consider an infinite-horizon MDP with a discounted sum-reward criterion as presented in \cite[Chap. 6]{Puterman1994}, defined by states, actions, transition probabilities, and rewards.
The  sum-reward criterion, called the gain, is computed as 
\begin{align}
G_t = \sum_{i=t}^\infty \lambda^{i-t} r_{i},
\end{align}
where $r_i$ is the reward received at each time step $i$ and $0 \leq \lambda<1$ the discount.

In general, the objective in this framework is to find a policy $\pi$, i.e., the decision rule to be used at each time step, that maximizes the expected gain
$\mathbb{E}[G_t]$. Here, the expectation is applied with respect to the randomness in the environment, modelled by transition probabilities between the system states.

Accordingly, the value of a state $s \in S$ obtained with a given policy $\pi$, where $S$ is the set of possible system states, is defined as the expected gain given that the system is in the state $s$ at time $t$:
\begin{align}
v^\pi(s) = \mathbb{E}[G_t|s].
\end{align}
Then, a policy $\pi^*$ is optimal if
\begin{align}
v^{\pi^*}(s) \geq v^\pi(s), \ \forall s \in S \text{ and } \forall \pi.
\end{align}
Under the optimal policy, the value of a state $s$ can be expressed via Bellman equation as
\begin{align}
v^{\pi^*}(s)  = \max_{a \in A_s} Q(s,a),
\end{align}
where $A_s$ is the set of allowable actions in state $s$ and  $Q(s,a)$, called the $Q$-value, is defined as
\begin{align}
\label{equ_Q}
Q(s,a) = \mathbb{E}[G_t|s,a] = R(s,a) + \lambda \sum_{s_j \in S} p(s_j|s,a) v^{\pi^*}(s_j),
\end{align}
with $p(s_j|s,a)$ denoting the transition probability from state $s$ to state $s_j$ given action $a$ and the average short-term reward $R(s,a)$ is $R(s,a) = \sum_j p(s_j|s,a) r(s_j,s)$, where $r(s_j,s)$ is the reward obtained when going from state $s$ to state $s_j$. Given the $Q$-values, the optimal action in a state $s$ is obtained as
\begin{align}
\label{eq_opt_act}
a^* = \underset{a \in A_{s}}{\text{ arg max } } Q(s,a).
\end{align}

A standard approach to compute the values $v^{\pi^*}(s)$ is to use the value iteration algorithm \cite[Chap 4.4]{Sutton2020}.
It consists in computing $v_{n+1}(s)$ at iteration $n+1$, $\forall s \in S$, as
\begin{align}
\label{equ_value_iteration}
v_{n+1}(s) = \max_{a \in A_s} \ \{ R(s,a) + \lambda \sum_{s_j \in S} p(s_j|s,a) v_{n}(s_j) \},
\end{align}
where $v_n(s)$ converges to $v^{\pi^*}(s)$ as it is the unique fixed point (see \cite{Puterman1994} for a proof). This step can be seen as a training phase.

Having the values of  $v^{\pi^*}(s)$ $\forall s \in S$, the action to be taken in a given state $s$ according to $\pi^*$ can be implemented by computing \eqref{equ_Q} $\forall a \in A_s$ and then using \eqref{eq_opt_act}.
We refer to the obtained policy as the maximum expected gain policy.

\section{Alternative proposed approach}

\subsection{Problem statement}

Instead of maximizing the expected gain, we propose to maximize the probability that $G_t$ is greater than a given value~$\alpha$. 
Since $p(G_t>\alpha)=1-p(G_t \leq \alpha)$, this is equivalent to minimizing $p(G_t<\alpha)$, commonly called outage probability.
This outage probability represents the risk to have a bad system outcome which is essential for safety applications.  
Therefore, the value of a state $s$ with parameter $\alpha$ is now defined as follows:
\begin{align}
v^{\pi}(s, \alpha) = p(G_t > \alpha |s).
\end{align}

Accordingly, we also define the value of a state with parameter $\alpha$ given an action $a \in A_{s}$:
\begin{align}
Q(s,a,\alpha) = p(G_t > \alpha|s,a).
\end{align}
The new goal is thus to find a policy $\pi^*$ that maximizes the probability that the gain $G_t$ is above an arbitrary value $\alpha$, i.e., such that:
\begin{align}
v^{\pi^*}(s, \alpha) \geq  v^{\pi}(s, \alpha), \ \forall s \in S \text{ and } \forall \pi.
\end{align}
This latter policy $\pi^*$ is called the alternative policy.

\subsection{Example}

We consider the recycling robot problem, presented in \cite[Chap. 3]{Sutton2020}, as a toy example to illustrate the difference between the two objectives. The problem is the following: 
A mobile robot running on a battery should collect empty cans. It can be in two battery states $S=\{s_1=\text{``low"}, s_2=\text{``high"}\}$. For both states, the possible actions are $A=\{ a_1=\text{``search", }a_2=\text{``wait" },a_3=\text{``recharge" }\}$. With the action ``wait", the robot stays in the same state with probability $p(s_i|s_i,a_2)=1$ (whatever $i$) and gets reward $r_{wait}$. With the action ``search", the robot goes from state ``high" to state ``low" with probability $p(s_1|s_2,a_1)=1-\beta$ and gets reward $r_{search}>r_{wait}$ with probability 1.  In the state ``low", it stays in the same state also with probability $\beta$ and gets reward $r_{search}$ in this case, but it gets the negative reward $r_{rescue}$ otherwise and goes back to state ``high". Finally, with the action ``recharge", the robot goes to state ``high" with probability $p(s_2|s_i,a_3)=1$ and gets reward~0.

For the simulations, we consider the following parameters: $r_{rescue}=-1$, $r_{wait}=0.4$, $r_{search}=0.9$, $\beta=0.8$, and the discount $\lambda=0.8$. In this case, the maximum expected gain policy consists in performing the action ``search" for both states, i.e., the robot always searches, regardless of the state. We also consider an alternative policy where the robot implements the action ``wait" in the state ``low".

\begin{figure}[h]
\centering
\includegraphics[width=0.8\columnwidth]{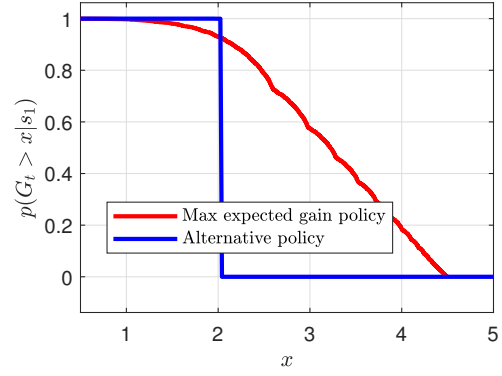}
\caption{Empirical CCDF, showing $p(G_t>x|s_1)$, obtained with the maximum expected gain policy and the alternative policy.}
\label{fig_CDF_1}
\end{figure}

With the alternative policy, the expected gain for state ``low" $\mathbb{E}[G_t|s_1]$ is reduced: 2 against 3.18.
However, $p(G_t>2-\epsilon|s_1)$, $\epsilon>0$ being a small quantity, is greater with the alternative policy: 
This can be observed on Figure~\ref{fig_CDF_1} where we show the empirical complement cumulative density function (CCDF) obtained by simulating both policies.
The maximum expected gain strategy is NOT optimal to maximize $p(G_t > \alpha|s_1)$ where $\alpha <2$.

\section{Connections with recent relevant works}
\label{sec_review_recent_work}

A relevant fundamental work from the literature is \cite{Bellemare2017}. 
In a nutshell, they propose to work directly with random variables, and thus distributions, rather than the expected gain. As a result, they consider the distributional Bellman equation as an alternative to the standard Bellman equation:
$Z(s,a) = R(s,a) + \lambda Z(s',a'),$
where $Z$ is the gain random variable (denoted by $G$  in this paper) given action $a$ in state $s$. 
This leads to a Monte Carlo algorithm (Algorithm~1 in the paper) where atoms and probabilities are tracked in the iteration loop. 

The objective in \cite{Bellemare2017} is not the outage probability but still the expected gain. 
Moreover, we focus on the MDP approach, meaning that we assume the knowledge of the model (i.e., the transition probabilities between the states). 
In \cite{Bellemare2017}, the practical part (Section~4) considers the Monte Carlo approach without the model. 

The authors of \cite{Bellemare2017} cover additional cases in a draft book currently under revision \cite{Bellemare2023}, including the practical distributional value iteration. As a result, our proposed Algorithm 1 is similar to the ``categorical dynamic programming" (Algorithm 5.3 in \cite[Chap. 5]{Bellemare2023}). 
Then, algorithm 2 can be classified as a ``risk-sensitive value iteration" (Section 7.7), where we use the outage probability\footnote{Only the conditional value at risk and the variance constrained objective are considered} rule as greedy policy. We note however that only temporal-difference learning with the Monte Carlo approach is considered, whereas  temporal-difference learning without Monte Carlo may be of interest, see Section~\ref{sec_app} and Appendix~\ref{sec_TD}.

\section{Proposed algorithm}

We first introduce an algorithm to compute $v(s,\alpha)=p(G_t > \alpha|s)$ under a given policy (Algorithm~\ref{first_alg}). 
This algorithm is then slightly modified to find a policy that maximizes $v(s,\alpha)$ (Algorithm~\ref{second_alg}). 

\subsection{Computing $v(s,\alpha)$ under a given policy}

\subsubsection{Notion of path to compute $v(s,\alpha)$}
\begin{figure}[h]
\centering
\includegraphics[width=1\columnwidth]{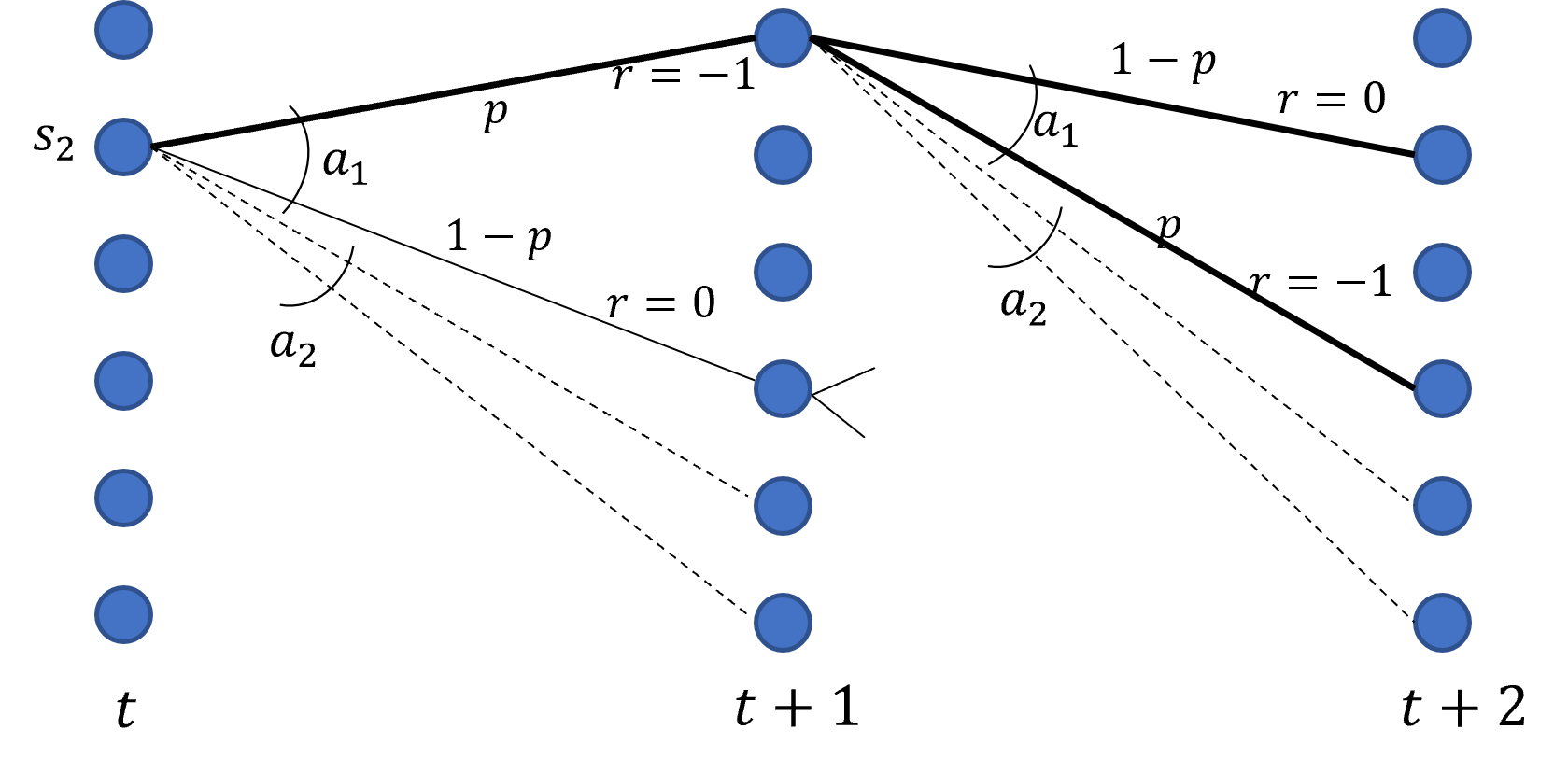}
\caption{Trellis representing the transitions between the states and the rewards at time $t$, $t+1$, and $t+2$. The labels on the edges show the state transition probabilities and short-term reward under several actions ($a_1$ and $a_2$).}
\label{fig_trellis_1}
\end{figure}

In this first subsection, we consider a given series of actions, e.g., established via the maximum expected gain policy.
Figure~\ref{fig_trellis_1} shows a trellis with the transitions between the states at time $t$, $t+1$, and $t+2$, of an arbitrary MDP.

We can define the notion of path in the trellis, corresponding to one possible set of successive (environment) realization starting from a given state. The $k$-th path starting from a state $s$ is defined by two values:
\begin{itemize}
\item A path probability: $P_{s}(k)$.
\item A path gain: $G_{s}(k)$.
\end{itemize} 
In the example of Figure~\ref{fig_trellis_1}, assume that the action $a_1$ is taken both at time $t$ and $t+1$ (illustrated by the plain lines).
The bold lines on the figure show two paths: a path with $P_{s_2}(1)= p \cdot (1-p)$ and $G_{s_2}(k)=-1 + \lambda \cdot 0$, a path with $P_{s_2}(2)= p \cdot p$ and $G_{s_2}(2)=-1 + \lambda \cdot (-1)$.

Let us merge two paths (i.e., add their probability values) if their gain difference is smaller than a some small value $\epsilon$.
Consequently, in the discounted infinite-horizon scope there is a finite number of paths with distinct values, say $K$.
We denote by 
\begin{align}
\label{eq_vec_proba_gain}
P_{s} = [P_{s} (1), ... , P_{s} (K)] \text{ and } G_{s} = [G_{s} (1), ...,G_{s} (K)]
\end{align}
the vectors representing the probabilities and the gains, respectively, for all paths starting from $s$. As a result, a state $s$ is characterized by the set $\Omega_{s}  = \{P_{s} , G_{s}  \}$. 
Then, $v(s,\alpha)$ can be computed as

\begin{align}
\label{eq_value_state_new}
v(s,\alpha) = \sum_{k=1}^K P_{s}(k) \cdot \mathbbm{1} \{ G_{s}(k) > \alpha \},
\end{align}
where $\mathbbm{1} \{\cdot\}$ denotes the indicator function.

\subsubsection{Recursively computing $v(s,\alpha)$}

\begin{figure}[h]
\centering
\includegraphics[width=0.7\columnwidth]{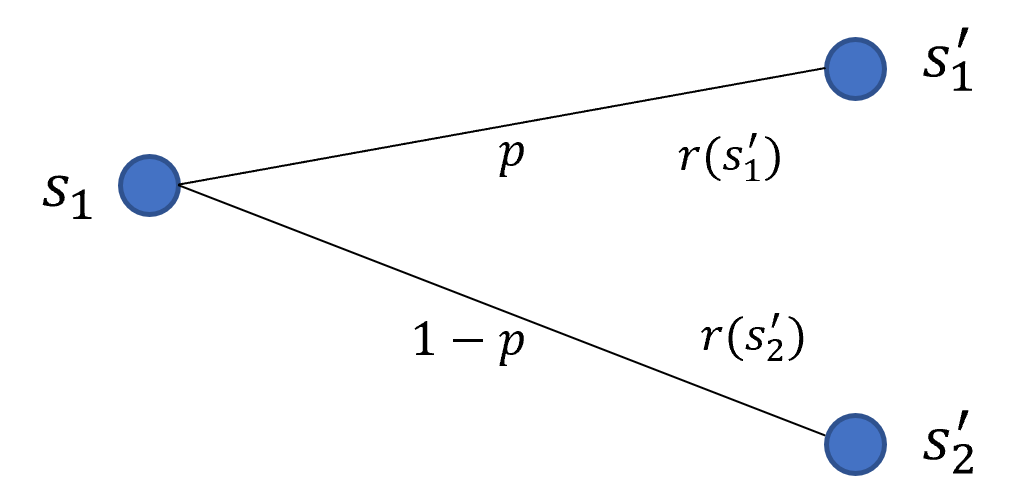}
\caption{One section of a trellis between a state $s_1$ and its subsequent states.}
\label{fig_trellis_2}
\end{figure}

Let us now consider one section of the trellis between a state $s_1$ and its subsequent states (assuming an arbitrary action): $s_1'$ with probability $p$ and reward\footnote{For the sake of simplicity we write $r(s_1')$ for $r(s_1,s'_1)$.} $r(s_1')$ and $s_2'$ with probability $1-p$ and reward $r(s_2')$, as illustrated on Figure~\ref{fig_trellis_2}.
Assume also that $\Omega_{s'_1} = \{P_{s'_1}, G_{s'_1} \}$ and $\Omega_{s'_2} = \{P_{s'_2}, G_{s'_2} \}$ are known and both vectors in $\Omega_{s'_1}$ and $\Omega_{s'_2}$ are of size $K$.

The set $\Omega_{s_1} = \{P_{s_1}, G_{s_1} \}$ can be computed from  $\Omega_{s'_1}$ and $\Omega_{s'_2}$ as follows:

\begin{align}
\label{eq_recursive_A}
&P_{s_1} = [p \cdot P_{s'_1}, (1-p) \cdot P_{s'_2})], \\
\label{eq_recursive_B}
&G_{s_1} = [r(s'_1) + \lambda \cdot G_{s'_1}, r(s'_2) + \lambda \cdot G_{s_2'}].
\end{align}

The main drawback of the above equation\footnote{The equation also brings out the main drawback of \eqref{eq_vec_proba_gain}: as the size doubles when moving from one trellis section to the previous one, $K$ is exponential in the trellis depth.} is that the size of the two resulting vectors is doubled compared to the ones of $\Omega_{s'_1}$ and $\Omega_{s'_2}$.

We propose to adopt a binning strategy to maintain the size of the vectors fixed. 
It is similar to the above rule where we merge two paths if their gain difference is smaller than some value $\epsilon$.
We introduce the vector $G^{ref}_{s}$ whose components  represent the center of the bins. 
The number of bins $K$ (which is different from the number of paths $K$ in \eqref{eq_vec_proba_gain}), their width, and their center is determined offline. For instance, the value of $G^{ref}_{s_1}(K/2)$ can be chosen as $v(s_1)$. Note that a unique vector $G^{ref}_{s}$, $\forall s  \in S$, could also be used to reduce the complexity.

Hence, after computing \eqref{eq_recursive_A} and \eqref{eq_recursive_B}, if two components of $G_{s_1}$ fall in the same bin of $G^{ref}_{s_1}$, e.g., the one centered at $G^{ref}_{s_1}(k)$, the corresponding probability values are added and the resulting value $P_{s_1}(k)$ is stored. To summarize, $G_{s_1}$ and $G_{s_1}^{ref}$ are  used to establish the binning rule, where $G_{s_1}$ is itself computed from $G^{ref}_{s'_1}$ and $G^{ref}_{s_2'}$ as (instead of \eqref{eq_recursive_B}) 

\begin{align}
\label{eq_recursive_B2}
G_{s_1} = [r(s'_1) + \lambda \cdot G^{ref}_{s'_1}, r(s'_2) + \lambda \cdot G^{ref}_{s_2'}].
\end{align}

As a result, we can use \eqref{eq_recursive_A} and \eqref{eq_recursive_B2}, as well as the binning trick, to iteratively approximate $\Omega_{s}$. This is summarized within Algorithm~\ref{first_alg}.

\begin{algorithm}
\caption{Computing $\Omega_{s}$ and $v(s,\alpha)$, $\forall s \in S$, with a given policy $\pi$. }
\label{first_alg}
\textbf{Input:}  $G^{ref}_{s}$ with $K$ elements, $\forall s \in~S$, a policy $\pi$, and $\alpha$. \\
\begin{algorithmic}[1]
\STATE Initialize $P_{s}$, $\forall s \in S$. 
\STATE For all $s\in S$, choose $a$ according to $\pi$ and compute
\begin{align}
\label{equ_sc_ref}
G_{s} = [r(s_{j_1}) + \lambda \cdot G^{ref}_{s_{j_1}}, ..., r(s_{j_k}) + \lambda \cdot G^{ref}_{s_{j_k}}, ... ],
\end{align}
for all states $s_{j_k}$ with a non-zero transition probability $p(s_{j_k}|s,a)$.
\WHILE{a stopping criterion is not met}
\FOR{all states $s \in S$}
\STATE Choose $a$ according to $\pi$ and compute 
\begin{align}
&P_{s} = [p(s_{j_1}|s,a) \cdot P_{s_{j_1}}, ... , p(s_{j_k}|s,a) \cdot P_{s_{j_k}}, ... ], 
\end{align}
for all states $s_{j_k}$ with a non-zero transition probability $p(s_{j_k}|s,a)$.
\STATE Reduce the size of the vector $P_{s}$ to $K$ elements via the binning of $G_{s}$ using $G_{s}^{ref}$: If two values of $G_{s}$ fall in the same bin, merge the corresponding probabilities by adding them.
\ENDFOR
\ENDWHILE
\STATE Compute $v(s,\alpha)$, $\forall s \in S$, using $\Omega_s$ via \eqref{eq_value_state_new}.
\STATE Return $\Omega_{s}$ and $v(s,\alpha)$, $\forall s \in S$.
\end{algorithmic}
\end{algorithm}

Since the binning rule is constant, the same probabilities of $P_s$ are merged at each iteration. Hence, steps 5 and 6 could be merged by directly computing a vector $P^{bin}_s$ of size $K$ where the probability values are added according to the binning rule.

\textbf{Example of binning rule:} On Figure~\ref{fig_trellis_2}, assume that $G_{s_1}^{ref} = G_{s'_1}^{ref} = G_{s'_2}^{ref}=[1, \ 3]$, $r(s'_1)=2$, $r(s'_2)=0$, and $\lambda=0.8$. Then, \eqref{eq_recursive_B2} yields $G_{s_1} = [2.8, \ 4.4, \ 0.8,\  2.4]$. Applying the binning rule yields $P^{bin}_{s_1}=[P_{s_1}(3), \ P_{s_1}(1)+P_{s_1}(2)+P_{s_1}(4)].$

Note that $\Omega_{s}$ can also be used to compute the CCDF, as an alternative to the empirical CCDF, without formerly running the algorithm.

\subsection{Finding a policy optimized for $p(G_t<\alpha)$}

In the previous subsection, the actions are chosen according to a given policy $\pi$, such as the maximum expected gain policy.  We now explain how to proceed to learn actions maximizing~$v(s,\alpha)$.

At steps 5-6 of Algorithm~\ref{first_alg},  $\Omega_{s}$ is computed for only one action $a$ determined by $\pi$. 
Alternatively, one could compute  $\Omega_{s,a}$ for all $a \in A_{s}$. With the obtained vectors at iteration $n$, one can compute
\begin{align}
\label{equ_alte_rule}
Q_n(s,a,\alpha) = \sum_k P_{s,a}(k) \cdot 1 \{ G^{ref}_{s}(k) > \alpha \}. 
\end{align}
Then, the action can be chosen as 
\begin{align}
\label{equ_alte_action}
a^* = \underset{a \in A_{s} }{\text{arg max }} Q_n(s,a,\alpha),
\end{align} 
and 
\begin{align}
v_n(s, \alpha)= Q_n(s,a^*,\alpha). 
\end{align}
As a result, Algorithm~\ref{first_alg} is modified to result in Algorithm~\ref{second_alg} to find a policy optimized for $p(G_t>\alpha)$.
Note that steps 5-6 of Algorithm~\ref{first_alg} are merged into a unique step 5 where the binning rule is directly applied to get the vector $P_{s,a}^{bin}$, as described at the end of the previous subsection.

The ouput of Algorithm~\ref{second_alg} (which can be seen as a training phase similarly to the value iteration algorithm) is $\Omega_s$. This can then be used in the inference phase to recover the optimal action to be performed at each state (as $v(s)$ is used to get the action  for the maximum expected gain policy, see the end of Section~\ref{sec_standard_expe_gain_policy}).

\begin{algorithm}
\caption{Finding a policy optimized for $p(G_t>\alpha)$. }
\label{second_alg}
\textbf{Input:}  $\alpha$ and $G^{ref}_{s}$ with $K$ elements, $\forall s \in S$. \\ 
\begin{algorithmic}[1]
\STATE Initialize $P_{s}$, $\forall s \in S$, and set $n=0$.
\STATE For all $s\in S$ and for all $a \in A_{s}$ compute
\begin{align}
G_{s,a} = [r(s_{j_1}) + \lambda \cdot G^{ref}_{s_{j_1}}, ..., r(s_{j_k}) + \lambda \cdot G^{ref}_{s_{j_k}}, ... ],
\end{align}
for all states $s_{j_k}$ with non-zero transition probabilities $p(s_{j_k}|s,a)$ and establish the binning rules accordingly.
\WHILE{a stopping criterion is not met}
\FOR{all states $s \in S$}
\STATE For all $a \in A_{s}$ compute 
\begin{align}
\label{eq_with_binning}
\begin{split}
&P^{bin}_{s,a} = [p(s_{j_1}|s,a) P_{s_{j_1}}(k_1)+p(s_{j_2}|s,a) P_{s_{j_2}}(k_2)+ \\
&..., ... ].
\end{split}
\end{align}
for all states $s_{j_k}$ with non-zero transition probabilities $p(s_{j_k}|s,a)$ and where the probabilities are added according to the binning rule of step 2.
\STATE For all $a \in A_{s}$ compute 
\begin{align}
\label{equ_q_value}
Q^n(s,a,\alpha) = \sum_k P^{bin}_{s,a}(k) \cdot 1 \{ G^{ref}_{s}(k) > \alpha \}, 
\end{align}
and set $a^* = \underset{a \in A_{s} }{\text{arg max }} Q(s,a,\alpha)$.
\STATE Set $P_{s}=P^{bin}_{s,a^*}$. 
 \ENDFOR
 \STATE Increment $n$.
\ENDWHILE
\STATE Return Return $\Omega_{s}$, $\forall s \in S$. 
\end{algorithmic}
\end{algorithm}

\subsection{Simulation results on the toy example}

Via Figure~\ref{fig_CDF_1} for the recycling robot example, we observe that the optimal policy for $p(G_t > \alpha |s_1)$ changes if $\alpha$ is greater or smaller than 2.
As an example, we run Algorithm~\ref{second_alg} with $\alpha=1.8$ and $\alpha=2.2$. In the first case, the CCDF of the policy obtained is the one shown with the blue curve on Figure~\ref{fig_CDF_1}. In the second case, we get the red curve. This yields the expected results, meaning that Algorithm~\ref{second_alg} manages to find the optimal strategy.

\section{Towards a deep learning extension}
\label{sec_toward_deep_learning}

\subsection{Temporal-difference learning}
\label{sec_TD}

A famous alternative to the value iteration algorithm is temporal-difference learning, see \cite[Chap.~6]{Sutton2020}, where the model is updated based on the difference between two estimates of a state. As an example, the $Q$ learning algorithm relies on this paradigm.
Temporal-difference learning also contains the main idea behind deep reinforcement learning algorithms, such as deep $Q$ learning \cite{Mnih2015}, as it enables to produce an error signal to train the neural networks.

Instead of looping over all the states to estimate the $v^{\pi^*}(s)$, the algorithm walks from state to state. 
In the standard implementation using the Monte Carlo approach, the system is in a state $s$ at time $t$ and an action $a$  is chosen by the algorithm. One then gets a reward $r(s,s')$ and the new state $s'$ of the system at time $t+1$. The estimate $\hat{Q}(s,a)$ of $Q^{\pi^*}(s,a)$ is updated as
\begin{align}
\hat{Q}(s,a) = \hat{Q}(s,a) + \gamma \Delta_t,
\end{align}
instead of \eqref{equ_value_iteration} with the value iteration algorithm. The quantity $\gamma$ is the learning rate and 
\begin{align}
\label{equ_error_signal_standard}
\Delta_t =\hat{Q}(s,a) - \hat{Q}'(s,a)
\end{align}
is the error signal, where:
\begin{itemize}
\item $\hat{Q}'(s,a) = r(s,s')+\lambda \max_{a'} \hat{Q}(s',a')$ is a first estimate of $Q^{\pi^*}(s,a)$ based on the observed reward and subsequent state obtained when taking the action $a$ in state~$s$.
\item $\hat{Q}(s,a)$ is a second estimate of $Q^{\pi^*}(s,a)$ obtained via the currently stored value.
\end{itemize}

Note that $\hat{Q}'(s,a)$ can also be computed via \eqref{equ_Q}, i.e., without the Monte Carlo approach but using the model.

In our framework, a similar error signal can be generated with the vector of probabilities, where we use temporal-difference idea but without the Monte Carlo approach as we assume knowledge of the model.
For a state $s$ and given a binning rule, $P_s$ can be updated as
\begin{align}
\label{equ_td_ext}
\begin{split}
&P_s = P_s + \gamma \Delta_t,
\end{split}
\end{align}
where $\Delta_t= f(P^{bin}_s, P_s)$ is the error signal with:
\begin{itemize}
\item $P^{bin}_s$ is obtained via \eqref{eq_with_binning} and \eqref{equ_q_value},
\item $P_s$ is the currently stored value.  
\end{itemize}
The function $f$ could be a distance measure between two distributions such as the KL divergence or the squared norm of the difference of the two vectors.
Of course, \eqref{equ_td_ext} needs to be normalized at each time step.

The drawback here, compared to standard temporal-difference learning relying on Monte Carlo method (and thus ``true" reinforcement learning), is that we still need a model of the environment dynamic (i.e., we need the transition probablities to compute \eqref{eq_with_binning}). 
For an algorithm compliant with the reinforcement learning paradigm, see Algorithm~1 in \cite{Bellemare2017} and Appendix~\ref{sec_app}.
Alternatively, a neural network could be trained in a Monte Carlo phase to infer the transition probabilities and then used within the algorithm described above.

\subsection{With neural networks}
\label{sec_neural_net_adapation}

The main idea behind deep $Q$ learning consists in using a neural network to compute $v(s)$ or $Q(s,a)$, $\forall s \in S$, and using the error signal $\Delta_t$ (or a sum of several error signals) to train the neural network.

We can propose a similar approach where a neural network computes $P_s$ and the error signal $\Delta_t =f(P^{bin}_s, P_s)$ is used for the training. 


\section{Conclusions}

In this paper, we introduced an algorithm to find a policy that maximizes the probability $p(G_t > \alpha)$ that the gain $G_t$ is higher than a predefined value $\alpha$. This is an alternative to algorithms searching for a policy that maximizes the expected gain.
Optimizing with respect to this new criterion $p(G_t > \alpha)$ induces computing path gains and probabilities, where a path corresponds to a given series of state transitions. Computing such path metrics is intractable as the number of paths is exponential in the depth of the trellis. 
As a result, we introduced a recursive calculation method and a binning rule to merge paths having similar gains. While this new algorithm is presented as an extension of the value iteration algorithm, the main principles are not restricted to this paradigm. We show in the last section how these principles can be applied to modify the $Q$ learning and deep $Q$ learning algorithms to optimize the alternative objective. We use temporal-difference learning but without the Monte Carlo approach. 

\section{Appendix}
\label{sec_app}
We discuss why the Monte Carlo approach seems to be efficient, as reported in \cite{Bellemare2017}.

Consider the example of Figure~\ref{fig_trellis_2}, with rewards $r(s'_1) = r(s'_2) = 0$.
Let $P_{s'_1}$ and $P_{s'_2}$ be the vector of probabilities at $s'_1$ and $s'_2$ respectively.
Let $P^*_{s_1} = p \cdot P_{s'_1} + (1-p) \cdot P_{s'_2}$ be the optimal probability vector to learn at $s_1$ via Monte Carlo samples  
and $P^{model}_{s_1}$ an estimation by a model.

In the non-Monte Carlo approach, we have access to $P^*_{s_1}$ to update $P^{model}_{s_1}$. The problem is trivial.
In the Monte Carlo approach, the error signal is computed based on a realization $(s,a,s'_i)$.
If $s'=s'_1$, the error signal is $f(P^{model}_{s_1},P_{s'_1})$, where $P_{s'_1}$ acts as the label.
If $s'=s'_2$, the error signal is $f(P^{model}_{s_1},P_{s'_2})$, where $P_{s'_2}$ acts as the label.
In the Monte Carlo training process, the learning algorithm gets approximately $p$ times the error $f(P^{model}_{s_1},P_{s'_1})$ and $(1-p)$ times the error $f(P^{model}_{s_1},P_{s'_2})$. Does it make $P^{model}_{s_1}$ converge to $P^*_{s_1}$?

We performed simple simulations, where $P_{s'_1}$ and $P_{s'_2}$ are chosen as discrete Gaussians with distinct means and where we train a neural network as follows.  We use the gradient of $f(P^{model}_{s_1},P_{s'_i})$ to update the model, based on a dataset containing $p$ labels $P_{s'_1}$ and $(1-p)$ label $P_{s'_2}$ (and where the input of the neural network is a constant). The output of the model $P^{model}_{s_1}$ converges towards $P^*_{s_1}$, which explains why the Monte Carlo approach is also efficient. 

It is still to be proven that there are no pathological case. Moreover, it takes several gradient steps to converge whereas the correct distribution is found with one step using the model.

\newpage


\begin{thebibliography}{99}

\bibitem{Bellemare2017} M. Bellemare, W. Dabney, and R. Munos, ``A distributional perspective on reinforcement learning," International conference on machine learning, pp. 449-458, July 2017. 

\bibitem{Bellemare2023} M. G. Bellemare, W. Dabney, and M. Rowland``Distributional Reinforcement Learning," MIT Press, http://www.distributional-rl.org, 2023.

\bibitem{Mnih2015}  Mnih et al., “Human-level control through deep reinforcement learning,” Nature, vol. 518, no. 7540, pp. 529–533, 2015.

\bibitem{Puterman1994} M. L. Puterman, ``Markov Decision processes: Discrete Stochastic Dynamic Programming," John Wiley $\&$ Sons, 1994.

\bibitem{Sibel2023} J.-C. Sibel, N. Gresset, and V. Corlay, ``An Application-oriented scheduler," 2023 IEEE Wireless Communications and Networking Conference (WCNC), March 2023. Online: https://arxiv.org/abs/2302.09926

\bibitem{Sutton2020} R. S. Sutton and A. G. Barto, ``Reinforcement Learning, An Introduction", 2nd ed., The MIT Press, 2020.


%
%
%
%
%
%
%
%
%
%
%
%
%
\end{thebibliography}
\end{document}